\documentclass{article}
\usepackage{color-edits}
\addauthor{sw}{blue}

\usepackage{microtype}
\usepackage{graphicx}
\usepackage{subfigure}
\usepackage{booktabs} % for professional tables
\usepackage{url}
\usepackage{amsmath,amssymb,amsfonts}
\usepackage{amsthm}
\usepackage{thmtools}
\usepackage{float}
\usepackage{pifont}

\usepackage{amsthm}
\usepackage{enumitem}

\usepackage{xcolor}
\definecolor{expert}{HTML}{008000}
\definecolor{error}{HTML}{f96565}

\usepackage{hyperref}

\usepackage[accepted]{icml2021}

\usepackage{tikz}
\usetikzlibrary{arrows,calc,bayesnet} 
\newcommand{\tikzAngleOfLine}{\tikz@AngleOfLine}
\def\tikz@AngleOfLine(#1)(#2)#3{%
\pgfmathanglebetweenpoints{%
\pgfpointanchor{#1}{center}}{%
\pgfpointanchor{#2}{center}}
\pgfmathsetmacro{#3}{\pgfmathresult}%
}

\newtheorem{theorem}{Theorem}

\declaretheoremstyle[
headfont=\normalfont\itshape,
qed=\qedsymbol,
]{mypf}
\declaretheorem[numbered=no, name=Proof, style=mypf]{pf}

\usepackage[font=small,labelfont=bf]{caption}
\def\BibTeX{{\rm B\kern-.05em{\sc i\kern-.025em b}\kern-.08em
    T\kern-.1667em\lower.7ex\hbox{E}\kern-.125emX}}

\usepackage[normalem]{ulem}

\icmltitlerunning{A Critique of Strictly Batch Imitation Learning}

\begin{document}

\twocolumn[
\icmltitle{A Critique of Strictly Batch Imitation Learning}



\icmlsetsymbol{equal}{*}

\begin{icmlauthorlist}
\icmlauthor{Gokul Swamy}{ri}
\icmlauthor{Sanjiban Choudhury}{aurora}
\icmlauthor{J. Andrew Bagnell}{aurora,ri}
\icmlauthor{Zhiwei Steven Wu}{isr}
\end{icmlauthorlist}

\icmlaffiliation{ri}{Robotics Institute, Carnegie Mellon University}
\icmlaffiliation{isr}{Institute for Software Research, Carnegie Mellon University}
\icmlaffiliation{aurora}{Aurora Innovation}

\icmlcorrespondingauthor{Gokul Swamy}{gswamy@cmu.edu}

\icmlkeywords{Machine Learning, ICML}

\vskip 0.3in
]



\printAffiliationsAndNotice{}  % leave blank if no need to mention equal contribution

\begin{abstract}
Recent work by Jarrett et al. attempts to frame the problem of offline imitation learning (IL) as one of learning a joint energy-based model \cite{jarrett2020strictly}, with the hope of out-performing standard behavioral cloning. We suggest that notational issues obscure how the psuedo-state visitation distribution the authors propose to optimize might be disconnected from the policy's \textit{true} state visitation distribution. We further construct natural examples where the parameter coupling advocated by Jarrett et al. leads to inconsistent estimates of the expert's policy, unlike behavioral cloning.
\end{abstract}

\section{Overview}
Consider a Markov Decision Process (MDP) parameterized by $\langle \mathcal{S}, \mathcal{A}, \mathcal{T}, r \rangle$, where $\mathcal{S}$ is the state space, $\mathcal{A}$ is the action space, $\mathcal{T}: \mathcal{S} \times \mathcal{A} \rightarrow \Delta(\mathcal{S})$ is the transition operator, $r: \mathcal{S} \times \mathcal{A} \rightarrow [-1, 1]$ is a reward function. For simplicity, we ignore the discount factor. We use $d_{\pi}(s, a)$ to refer to the state-action visitation distribution of policy $\pi_{\theta}: \mathcal{S} \rightarrow \Delta(\mathcal{A})$. 

We focus on a recently proposed method for offline imitation learning: Energy-based Distribution Matching (EDM) \cite{jarrett2020strictly}. The authors propose to minimize the KL divergence from the demonstrator's state-action marginal distribution ($d_{\pi_E}$) to some state-action distribution $p_{\pi_{\theta}}$:
\begin{align}
  J(\pi_{\theta}) &= \mathop{{}\mathbb{E}}_{s, a \sim d_{\pi_E}}[- \log p_{\pi_{\theta}}(s, a)] \\
  &= \mathop{{}\mathbb{E}}_{s, a \sim d_{\pi_E}}[-\log \pi_{\theta}(a|s) - \log p_{\theta}(s)] \label{eq:obj}
\end{align}
We find two key issues with the proposed approach:
\vspace{-1em}
\begin{enumerate}[noitemsep,topsep=5pt]
    \item The function $J(\pi_\theta)$ does not properly capture the divergence from the demonstrator since the second term in \eqref{eq:obj} might not have anything to do with the policy's true state visitation distribution.
    \item As it uses a single set of parameters to represent both the learner's policy and state visitation distribution, the authors' method might not recover the expert's policy.
\end{enumerate}
\vspace{-1em}
We now provide evidence to substantiate our concerns.
\section{Notational Concerns}
As indicated in \eqref{eq:obj}, the authors derive a policy from $p_{\theta}(s, a)$ by conditioning on state. While this is a valid mathematical operation, unless one explicitly enforces $p_{\pi_{\theta}}(s, a) = d_{\pi}(s, a)$, $p_{\theta}(s)$ might not have anything to do with the actual state visitation distribution of the policy, $d_{\pi}(s)$. In particular, even though $p_{\pi_{\theta}}(s, a) = \pi_{\theta}(a|s) p_{\theta}(s)$ and $d_{\pi}(s, a) = \pi_{\theta}(a|s) d_{\pi}(s)$, we cannot say that $p_{\theta}(s) = d_{\pi}(s)$ without further information. While computing $d_{\pi}(s)$ requires samples from the environment, $p_{\theta}(s)$ is purely a function of parameters $\theta$ and can be set completely independently of the policy's true state visitation distribution. We now provide a concrete example of such a discrepancy

\noindent\textbf{Example 1. } Consider an MDP with a fixed initial state $s_0$ and no transitions between different states, only self-loops. Then, $d_{\pi}(s)$ is a singleton distribution with all mass on $s_0$. We can choose $p_{\theta}(s, a) = \frac{1}{|S||A|}$. Then, $p_{\pi_{\theta}}(s) = \sum_a \frac{1}{|S||A|} = \frac{1}{|S|}$, which differs at every state from $d_{\pi}(s)$.\hfill\qedsymbol

Contrary to the authors' claim that ``backpropagating through the first term is impossible as we cannot
compute $p_{\theta}(s)$," one can indeed compute $p_{\theta}(s)$ by simply summing or integrating out actions. What is much more challenging to do is computing $d_{\pi}(s)$ without access to the environment. However, as established by the above example, there is no apriori reason to connect $p_{\theta}(s)$ with $d_{\pi}(s)$.

We now detail how these concerns apply for the particular parameterization the authors assume. Notice that the first term of \eqref{eq:obj} is the behavioral cloning (BC) objective. To evaluate the latter term, the authors fix the policy to be a member of the exponential family:
\begin{equation}
    \pi_{\theta}(a|s) = \frac{e^{f_{\theta}(s)[a]}}{\sum_a e^{f_{\theta}(s)[a]}}
\end{equation}
\clearpage
and proceed by constructing an ``energy-based'' model:
\begin{align}
  E_{\theta}(s) = \log \sum_a e^{f_{\theta}(s)[a]} \\
  p_{\theta}(s) = \frac{e^{-E_{\theta}(s)}}{\sum_s e^{-E_{\theta}(s)}} \label{eq:svd}
\end{align}
They sample from this distribution using the Langevin method. Notice that as in Example 1, the transition probabilities of the MDP do not appear anywhere in this expression, which is suggestive of a concern. The authors proceed to note that the exponential family is invariant to an additive shift applied to each $f_{\theta}(s)[\cdot]$ score. Thus, they argue, one could use these additive shifts to capture the state visitation distribution of the policy. While there does exist a state-dependent additive shift that captures the true state visitation distribution of the policy (because we could use one shift per state), there is no guarantee -- and seemingly no connection --  between the state distribution in \eqref{eq:svd} and that of the policy. More explicitly, for any choice of $g(s)$, the following parameterization captures the same set of policies the authors assume:
\begin{align}
  \pi_{\theta}(a|s) = \frac{e^{f_{\theta}(s)[a] + g(s)}}{\sum_a e^{f_{\theta}(s)[a] + g(s)}} 
%   \text{s.t. } \forall s \in \mathcal{S}, \sum_a f_{\theta}(s)[a] = 0.
\end{align}
Recall that because the softmax operator is invariant to an additive shift to all logits like $g(s)$, the policy's true state visitation distribution is dependent \textit{only} on $f_{\theta}$. However, it would have the following sampling distribution:
\begin{equation}
    p_{\theta}(s) = \frac{e^{-g(s)} / \sum_a e^{f_{\theta}(s)[a]}}{\sum_s (e^{-g(s)} / \sum_a e^{f_{\theta}(s)[a]})}
\end{equation}
While changing $g(s)$ does not affect the policy (and thus, $d_{\pi}(s)$) whatsoever, different settings of $g(s)$ can lead to very different values of $p_{\theta}(s)$. We now provide an example where ${p}_{\theta}(s) \neq d_{\pi}(s)$.

\noindent\textbf{Example 2. } Consider setting $g(s) = -\log \sum_a e^{f_{\theta}(s)[a]}$. Then,
\begin{align}
{p}_{\theta}(s) &= \frac{e^{ \log \sum_a e^{f_{\theta}(s)[a]}} / \sum_a e^{f_{\theta}(s)[a]}}{\sum_s (e^{\log \sum_a e^{f_{\theta}(s)[a]}} / \sum_a e^{f_{\theta}(s)[a]})} \\
& = \frac{1}{\sum_s 1}
\end{align}
This is a uniform distribution over state space $U(\mathcal{S})$, regardless of the policy. As in Example 1, we can easily select policy parameters $\theta$ for any MDP without trivial transition dynamics such that $d_{\pi}(s) \neq U(\mathcal{S})$.  \hfill \qedsymbol

Thus, even in the particular parameterization assumed by the authors, $p_{\theta}(s)$ might be entirely disconnected from the policy's true state visitation distribution. We thank the authors for, in response to this note, publishing a \href{https://www.vanderschaar-lab.com/clarification-on-strictly-batch-imitation-learning-by-energy-based-distribution-matching/}{blog post} recognizing the notational issue in their original paper. 

\section{Inconsistency Concerns}
We now detail how using the same set of parameters to represent both the policy and ${p}_{\theta}(s)$ (Desiderata 2 in \citet{jarrett2020strictly}) can lead to a gradient that points away from the expert's policy, even in the limit of infinite demonstrations.

\noindent\textbf{Example 3.} 
Consider an MDP with arbitrary dynamics and two states, $s_1$ and $s_2$, each of which has two actions, $a_1$ and $a_2$. Assume that both the learner and expert have the following policy class:
\begin{equation}
   \pi_{\theta}(a_2|s_1) = \frac{e^{\theta}}{1 + e^{\theta}}\text{, }\pi_{\theta}(a_2|s_2) = \frac{e^{k\theta}}{1 + e^{k\theta}} 
\end{equation}
Note that this means that there exists a policy in the learner's policy class that can exactly match the expert. Initialize $\theta = 0$ and suppose that the expert policy has $\theta_E = 1$. $\mathop{{}\mathbb{E}}_{(s, a) \sim \mathcal{D}_E}[\nabla_{\theta} \log \pi_{\theta}] > 0$, which points towards the expert's policy. Then, considering the second term in the authors' objective:
\begin{align}
\log {p}_{\theta}(s_2) &= \log(1 + e^{k\theta}) - \log(1 + e^{\theta} + 1 + e^{k\theta}) \\
\nabla_{\theta} \log {p}_{\theta}(s_2) &= \frac{ke^{k\theta}}{1 + e^{k\theta}} - \frac{e^{\theta} + ke^{k\theta}}{2 + e^{\theta} + e^{k\theta}}
  \end{align}
We can plug in our initial condition of $\theta = 0$ to arrive at:
\begin{equation}
    \nabla_{\theta} \log{p}_{\theta}(s_2)|_{\theta=0} = \frac{k}{2} - \frac{1 + k}{4} = \frac{k - 1}{4}
\end{equation}
Thus, by setting $k = \frac{1}{2} < 1$, $\nabla_{\theta} \log {p}_{\theta}(s_2) < 0$, which pushes the parameter away from the expert's policy. \hfill \qedsymbol

Moreover, the gradient of the second term in \eqref{eq:obj} can be nonzero at the expert's policy, pushing a learner that is already perfectly mimicking the expert away from the expert's policy.
\begin{theorem}
There exists an MDP and a realizable expert policy such that the gradient of objective \eqref{eq:obj} is nonzero at the expert's policy.
\end{theorem}

\begin{pf}
 Consider the same setup as Example 3. Then, $\nabla_{\theta} \log p_{\theta}(s_2)|_{\theta=1} \approx -0.245 \neq 0$. Because we are already at the expert's policy, $\mathop{{}\mathbb{E}}_{(s, a) \sim \mathcal{D}_E}[\nabla_{\theta} \log \pi_{\theta}]|_{\theta=1}= 0$. Thus, the overall gradient of \eqref{eq:obj} would be nonzero.
\end{pf}

Intuitively, for $0 < k < 1$, the expert prefers $a_2$ in both states, but prefers it \textit{more} in $s_1$. However, without knowing anything about the transition dynamics of the MDP, one has no idea how this relative preference translates to a state visitation distribution. Theorem 1 shows that optimizing \eqref{eq:obj} can sometimes amount to moving along an arbitrary direction in parameter space that might lead away from the expert's policy. Thus, the authors' procedure can lead to inconsistent estimates of the expert's policy, even with infinite expert demonstrations, unlike behavioral cloning.

\bibliography{example_paper}
\bibliographystyle{icml2021}

\end{document}